\documentclass{article}

     \usepackage[nonatbib,preprint]{neurips_2019}

\usepackage{microtype}
\usepackage{graphicx}
\usepackage{booktabs} 
\usepackage[lofdepth,lotdepth]{subfig}
\usepackage[font=footnotesize,labelfont=bf]{caption}
\usepackage{algorithm}
\usepackage{algorithmic}

\usepackage[utf8]{inputenc} 
\usepackage[T1]{fontenc}    
\usepackage{hyperref}       
\usepackage{url}            
\usepackage{amsfonts}       
\usepackage{nicefrac}       

\usepackage{hyperref}
\usepackage{authblk}
\usepackage{enumitem}

\title{CapsAttacks: Robust and Imperceptible Adversarial
Attacks on Capsule Networks}

\author[1]{Alberto Marchisio}
\author[2]{Giorgio Nanfa}
\author[1]{Faiq Khalid}
\author[1]{Muhammad Abdullah Hanif}
\author[2]{\\Maurizio Martina}
\author[1]{Muhammad Shafique}

\affil[1]{Technische Universität Wien (TU Wien), Vienna, Austria}
\affil[2]{Politecnico di Torino (PoliTo), Turin, Italy}
\affil[ ]{Email: \{alberto.marchisio, faiq.khalid, muhammad.hanif, muhammad.shafique\}@tuwien.ac.at}
\affil[ ]{giorgio.nanfa@studenti.polito.it, maurizio.martina,@polito.it}

\begin{document}

\maketitle

\begin{abstract}
Capsule Networks preserve the hierarchical spatial relationships between objects, and thereby bears a potential to surpass the performance of traditional Convolutional Neural Networks (CNNs) in performing tasks like image classification. A large body of work has explored adversarial examples for CNNs, but their effectiveness on Capsule Networks has not yet been well studied. In our work, we perform an analysis to study the vulnerabilities in Capsule Networks to adversarial attacks. These perturbations, added to the test inputs, are small and imperceptible to humans, but can fool the network to mispredict. We propose a greedy algorithm to automatically generate targeted imperceptible adversarial examples in a black-box attack scenario. We show that this kind of attacks, when applied to the German Traffic Sign Recognition Benchmark (GTSRB), mislead Capsule Networks. Moreover, we apply the same kind of adversarial attacks to a 5-layer CNN and a 9-layer CNN, and analyze the outcome, compared to the Capsule Networks to study differences in their behavior.
\end{abstract}
\section{Introduction}
\label{introduction}

Convolutional Neural Networks (CNNs) showed a great improvement and success in many Machine Learning (ML) applications, e.g., object detection, face recognition, image classification (Bhandare et al., 2016). However, the spatial hierarchies between the objects, e.g., orientation, position and scaling, are not preserved by the convolutional layers. CNNs are specialized to identify and recognize the presence of an object as a feature, without taking into account the spatial relationships across multiple features. Recently, Sabour and Hinton et al. (2017) proposed the CapsuleNet, an advanced Neural Network architecture composed of so-called capsules, which is trained based on the Dynamic Routing algorithm between capsules. The key idea behind the CapsuleNet is called inverse graphics: when the eyes analyze an object, the spatial relationships between its parts are decoded and matched with the representation of the same object in our brain. Similarly, in CapsuleNets the feature representations are stored inside the capsules in a vector form, in contrast to the scalar form used by the neurons in traditional CNNs (Mukhometzianov and Carrillo, 2018). Despite the great success in the field of image classification, recent works (Yoon, 2017) have demonstrated that, in a similar way as CNNs, the CapsuleNet is also not immune to adversarial attacks. Adversarial examples are small perturbations added to the inputs, which are generated for the purpose to mislead the network. Since these examples can fool the network and reveal the corresponding security vulnerabilities, they can be dangerous in safety-critical applications, like Voice Controllable Systems (VCS) (Carlini et al, 2016; Zhang et al., 2017) and traffic signs recognition (Kukarin et al., 2017; Yuan et al., 2017). Many works (Hafahi et al., 2018; Kukarin et al., 2017) have analyzed the impact of adversarial examples in CNNs, and studied different methodologies to improve the defense solutions. Towards CapsuleNets, in this paper, we aim to addressing the following \textbf{fundamental research questions:}
\begin{enumerate}[leftmargin=*]
\vspace*{-2mm}
    \item Is a CapsuleNet vulnerable to adversarial examples?
    \vspace*{-1mm}
    \item How does the CapsuleNet's vulnerability to adversarial attacks differ from that of the traditional CNNs?
    \vspace*{-2mm}
\end{enumerate}
To the best of our knowledge, we are the first to study of the vulnerability of the CapsuleNet to such adversarial attacks for the German Traffic Sign Recognition Benchmark - GTSRB (Houben et al., 2013), which is crucial for autonomous vehicle use cases. Moreover, we compare the CapsuleNet to CNNs with 5 and 9 layers, by applying affine transformations and adversarial attacks.

\textbf{Our Novel Contributions:}
\begin{enumerate}[leftmargin=*]
\vspace*{-2mm}
\item We analyze the behavior of the CapsuleNet, compared to a 5-layer CNN and a 9-layer CNN, under affine transformations applied to the input images of the GTSRB dataset (see \textbf{Section 2}).
\vspace*{-1mm}
\item We develop a novel algorithm to automatically generate targeted imperceptible and robust adversarial examples (see \textbf{Section 3}).
\vspace*{-1mm}
\item We compare the robustness of the CapsuleNet with a 5-layer CNN and a 9-layer CNN, under the adversarial examples generated by our algorithm (see \textbf{Section 4}).
\vspace*{-2mm}
\end{enumerate}

Before  proceeding  to  the  technical  sections,  we  present  an overview  of  the  CapsuleNets  (in \textbf{Section  1.1}) and of the adversarial attacks (in \textbf{Section 1.2}),  to  a  level  of  detail necessary to understand the contributions of this paper.


\subsection{Background: CapsuleNets}

\begin{figure}[t]
\noindent
\begin{minipage}[t]{.49\linewidth}
\begin{figure}[H]
	\centering
	\vspace*{5mm}
	\includegraphics[width=\linewidth]{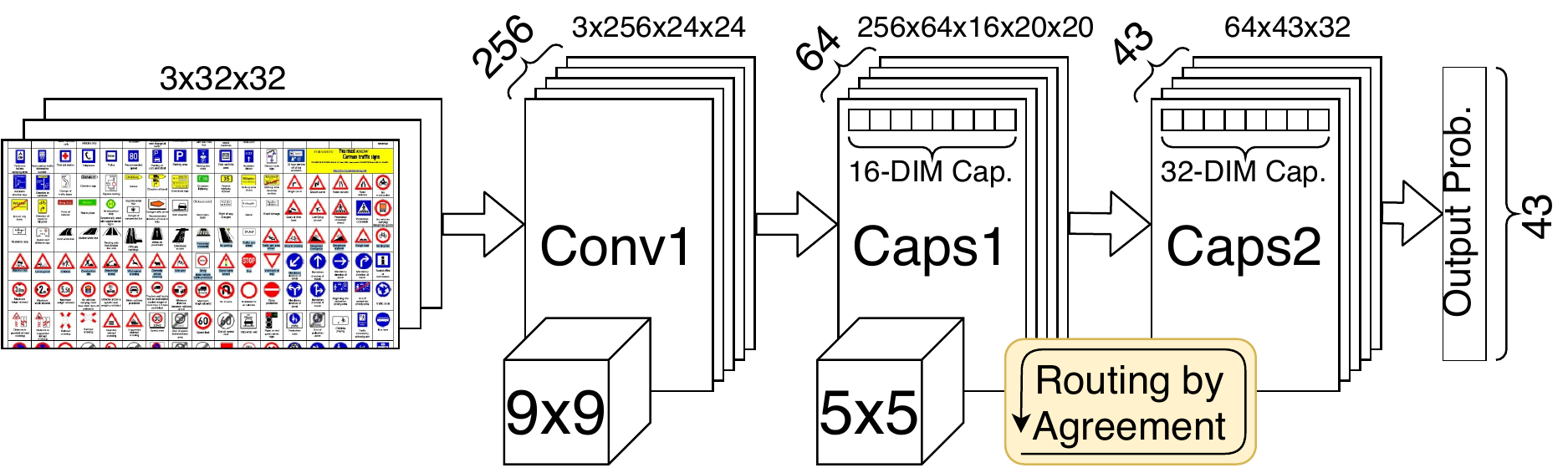}
	\caption{Architecture of the CapsuleNet for the\\GTSRB dataset.}
	\label{fig:capsnet_GTSRB}
\end{figure}
\end{minipage}
\hfill
\begin{minipage}[t]{.49\linewidth}
\begin{figure}[H]
	\centering
	\includegraphics[width=\linewidth]{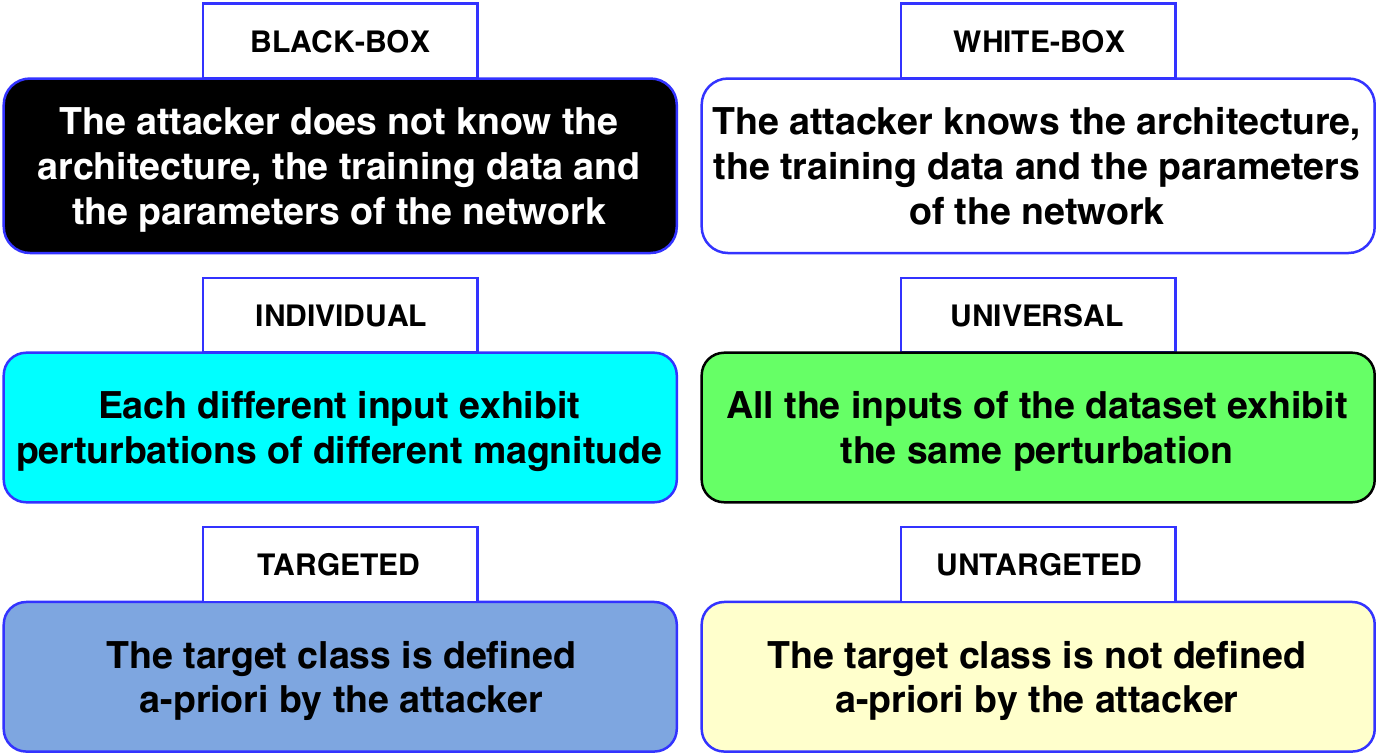}
	\caption{Taxonomy of Adversarial Examples.}
	\label{fig:att_types}
\end{figure}
\end{minipage}
\end{figure}

Capsules were first introduced by Hinton et al. (2011). They are multi-dimensional entities that are able to learn hierarchical information of the features. Compared to traditional CNNs, a CapsuleNet has the \textit{capsule} (i.e., a group of neurons) as the basic element, instead of the neuron. Recent works about CapsuleNet's architecture and training algorithms (Hinton et al., 2018; Sabour et al., 2017) have shown competitive accuracy results for image classification task, compared to other state-of-the-art classifiers. Kumar et al. (2018) proposed a CapsuleNet architecture, composed of 3 layers, which achieves good performance for the GTSRB dataset (Houben et al., 2013). The architecture is shown in Figure \ref{fig:capsnet_GTSRB}. Note, between the two consecutive capsule layers (i.e., Caps1 and Caps2), the routing-by-agreement algorithm is performed. It is demanding from a computational point of view because it introduces a feedback loop during the inference.

\subsection{Background: Adversarial Attacks}

\begin{figure}[t]
\centering
\begin{minipage}[t]{.47\linewidth}
\subfloat[]{
\includegraphics[width=\linewidth]{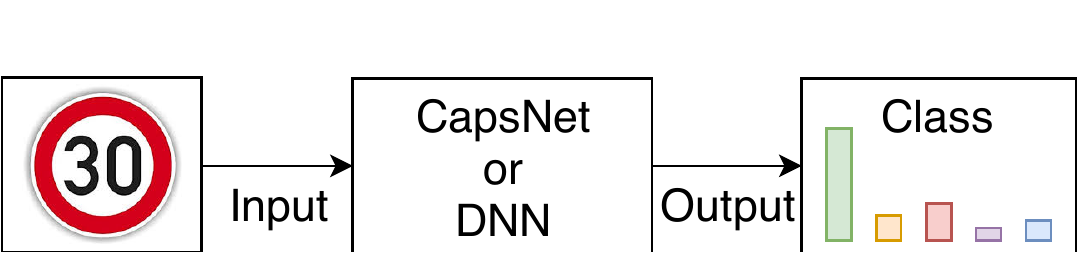}
\label{fig:capsattack_overview_clean}}
\end{minipage}
\hfill
\begin{minipage}[t]{.47\linewidth}
\subfloat[]{
\includegraphics[width=\linewidth]{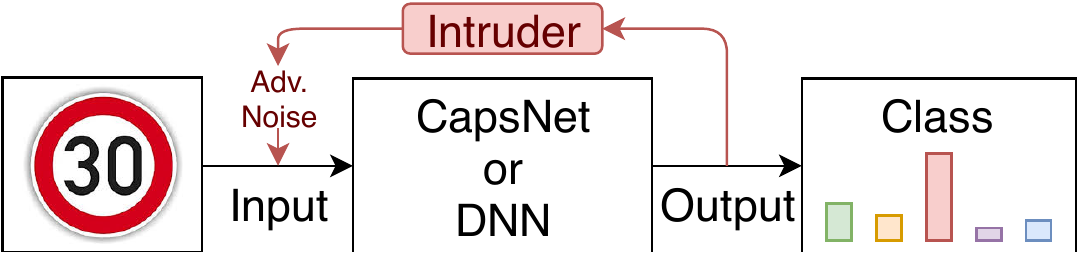}
\label{fig:capsattack_overview_intruder}}
\end{minipage}
\caption{An adversarial attack under black-box assumption. (a) Classification for a clean image.\\(b) Classification for an image perturbed by an adversarial intruder.}
\label{fig:capsattack_overview}
\end{figure}

Szegedy et al. (2014) studied that several machine learning models are vulnerable to adversarial examples. Goodfellow et al. (2015) explained the problem observing that \textit{machine learning models misclassify examples that are only slightly different from correctly classified examples drawn from the data distribution}. Considering an input example $x$, the adversarial example $x^*=x+\eta$ is equal to the original one, except for a small perturbation $ \eta $. The goal of the perturbation is to maximize the prediction error, in order to make the predicted class $C(x)$ different from the target one $C(x^*)$. In recent years, many methodologies to generate adversarial examples and their respective defense strategies have been proposed (Feinman et al., 2017; Bhagoji et al., 2017a; Bhagoji et al., 2017b). Adversarial attacks can be categorized according to different attributies, e.g., the choice of the class, the kind of the perturbation and the knowledge of the network under attack (Papernot et al., 2017; Yuan et al., 2017). We summarize these attributies in the Figure \ref{fig:att_types}. A simple representation of the attack scenario that we consider throughout this paper is visible in Figure \ref{fig:capsattack_overview}. An adversarial attack is very efficient if it is \textit{imperceptible} and \textit{robust}: this is the main concept of the analysis conducted by Luo et al. (2018). Their attack modified the pixels in high variance areas, since the human eyes do not perceive their modifications much. Moreover, an adversarial example is robust if the gap between the probabilities of the predicted and the target class is so large that, after an image transformation (e.g., compression or resizing), the misclassification still holds.

Recent works showed that a CapsuleNet is vulnerable to adversarial attacks. Jaesik Yoon (2017) analyzed the CapsuleNet's accuracy applying Fast Gradient Sign Method (FGSM), Basic Iteration Method (BIM), Least-likely Class Method and Iterative Least-likely Class Method (Kukarin et al., 2017) to the MNIST dataset (LeCunn et al., 1998). Frosst et al. (2017) presented an efficient technique to detect the crafted images on  MNIST, Fashion-MNIST (Xiao et al., 2017) and SVHN (Netzer et al., 2011) datasets.

\section{Analysis: Affine Transforming the Images}
\label{sec:affine_transformation}

Before studying the vulnerability of the CapsuleNet under adversarial examples, we apply some affine transformations to the test input images of the GTSRB dataset and observe their effects on our network predictions. We use three different types of transformations: rotation, shift and zoom. This analysis is important to understand how affine transformations, which are perceptible yet plausible in the real world, can or cannot mislead the networks under investigation.

\subsection{Experimental Setup}

We consider the architecture of the CapsuleNet, as shown in Figure \ref{fig:capsnet_GTSRB}. It is composed of a convolutional layer, with kernel $9 \times 9$, a convolutional capsule layer, with kernel $5 \times 5$, and a fully connected capsule layer. We implement it in TensorFlow, to perform classification on the GTSRB dataset (Kumar et al., 2018). This dataset has images of size $32 \times 32$ and it is divided into 34799 training examples and 12630 testing examples. Each pixel intensity assumes a value from 0 to 1. The number of classes is 43. For evaluation purposes, we compare the CapsuleNet with a 5-layer CNN (LeNet) (Ameen, 2018), trained for 30 epochs, and a 9-layer CNN (VGGNet) (Ameen, 2018), trained for 120 epochs. They are both implemented in TensorFlow and their accuracy with clean test images are 91.3\% and 97.7\%, respectively.

\subsection{Robustness under Affine Transformations}
\label{subsec:affine}

\begin{figure}[t]
\centering
\subfloat[]{
\includegraphics[width=.95\linewidth]{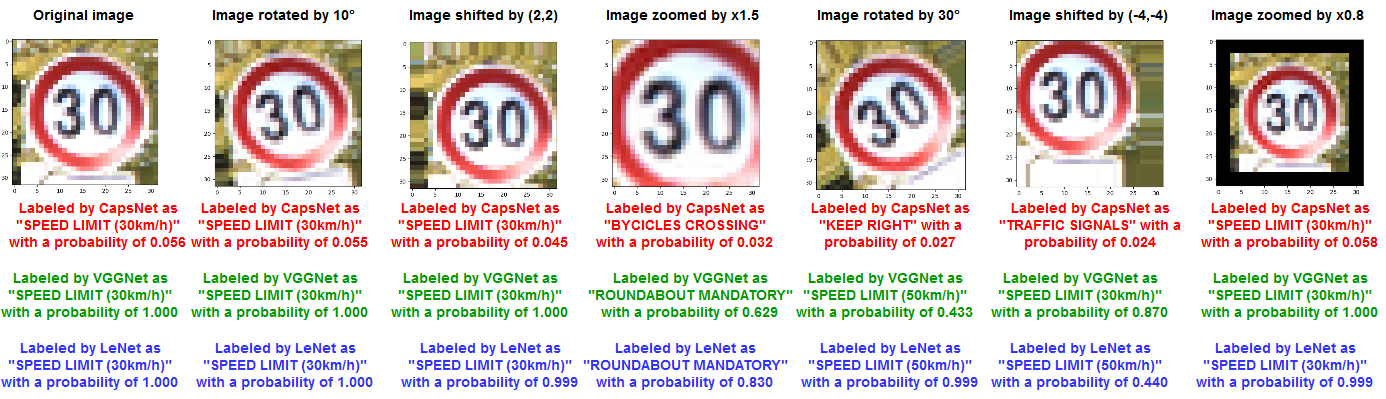}
\label{fig:30affine}}\\
\subfloat[]{
\includegraphics[width=.95\linewidth]{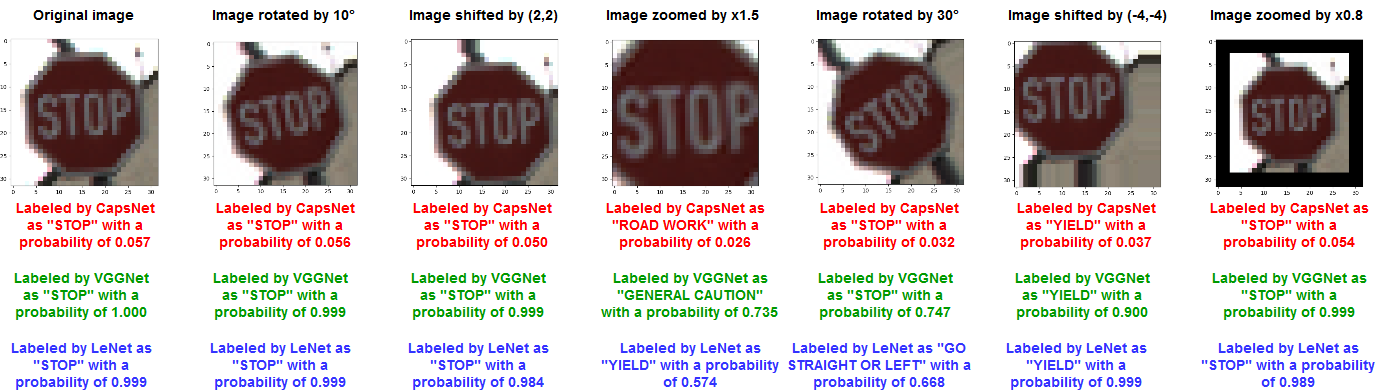}
\label{fig:stopaffine}}
\caption{Affine transformations on the test images, with the corresponding classification predictions made by the CapsuleNet, the VGGNet and the LeNet. (a) Example of a ``30 km/h speed limit'' sign. (b) Example of a ``Stop'' sign.}
\label{fig:affinetransformations}
\end{figure}

Some examples of affine transformations applied to the input are shown in Figure \ref{fig:affinetransformations}. The analysis shows that both the CapsuleNet and the VGGNet can be fooled by some affine transformations, like zoom or shift. while the confidence of the CapsuleNet is lower. Moreover, the LeNet, since it has lower number of layers, compared to the VGGNet, is more vulnerable to this kind of transformations, as we expected. The CapsuleNet, with capsules and advanced algorithms, such has the so-called routing-by-agreement, is able to overcome the lower complexity, in terms of number of layers and parameters, compared to the VGGNet. Indeed, as we can notice in the example of the image rotated by 30$^{\circ}$ of Figure \ref{fig:stopaffine}, the confidence is lower, but both the CapsuleNet and the VGGNet are able to classify correctly, while the LeNet is fooled.

\section{Generation of
Targeted Imperceptible and Robust Adversarial
Examples}
\label{sec:attack}

An efficient adversarial attack can generate imperceptible and robust examples to fool the network. Before describing the details of our algorithm, we discuss the importance of these two concepts.

\subsection{Imperceptibility and Robustness}

An adversarial example can be defined \textit{imperceptible} if the modifications of the original sample are so small that humans cannot notice them. To create an imperceptible adversarial example, we need to add the perturbations in the pixels of the image with the highest standard deviation. In fact the perturbations added in high variance zones are less evident and more difficult to detect with respect to the ones applied in low variance pixels. Considering an area of $M \cdot N$ pixels \textit{x}, the standard deviation ($SD$, Equation \ref{eq:SD}) of the pixel $x_{i,j}$ can be computed as the square root of the variance, where $\mu$ is the average of the $M \cdot N$ pixels:
\begin{equation}
    SD(x_{i,j})=\sqrt{\frac{ \sum\limits_{k=1}^M \sum\limits_{l=1}^N (x_{k,l}-\mu)^2-(x_{i,j}-\mu)^2}{M \cdot N}}
    \label{eq:SD}
\end{equation} 
Hence, when the pixel is in a high variance region, its standard deviation is high and the probability to detect a modification of the pixel is low. To measure the imperceptibility, it is possible to define the distance ($D$, Equation \ref{eq:distance}) between the original sample X and the adversarial sample X*, where $\delta_{i,j}$ is the perturbation added to the pixel $x_{i,j}$:
\begin{equation}
    D(X^*,X)=\sum\limits_{i=1}^{M} \sum\limits_{j=1}^{N}\frac{\delta_{i,j}}{SD(x_{i,j})}
    \label{eq:distance}
\end{equation} 
This value indicates the total perturbation added to all the pixels under consideration. We define also $D_{MAX}$ as the maximum total perturbation tolerated by the human eye.

An adversarial example can be defined \textit{robust} if the gap function, i.e., the difference between the probability of the target class and the maximum class probability is maximized:
\begin{equation}
    GAP=P(target \,  class)-max\{P(other \,  classes)\}
    \label{eq:gap}
\end{equation} 
If the gap function increases, the adversarial example becomes more robust, because the modifications of the probabilities caused by some image transformations (e.g., compression or resizing) tend to be less effective. Indeed, if the gap function is high, a variation of the probabilities could not be sufficient to achieve a misclassification.

\subsection{Generation of the Adversarial Examples}

\begin{figure}[t]
\centering
\includegraphics[width=.9\linewidth]{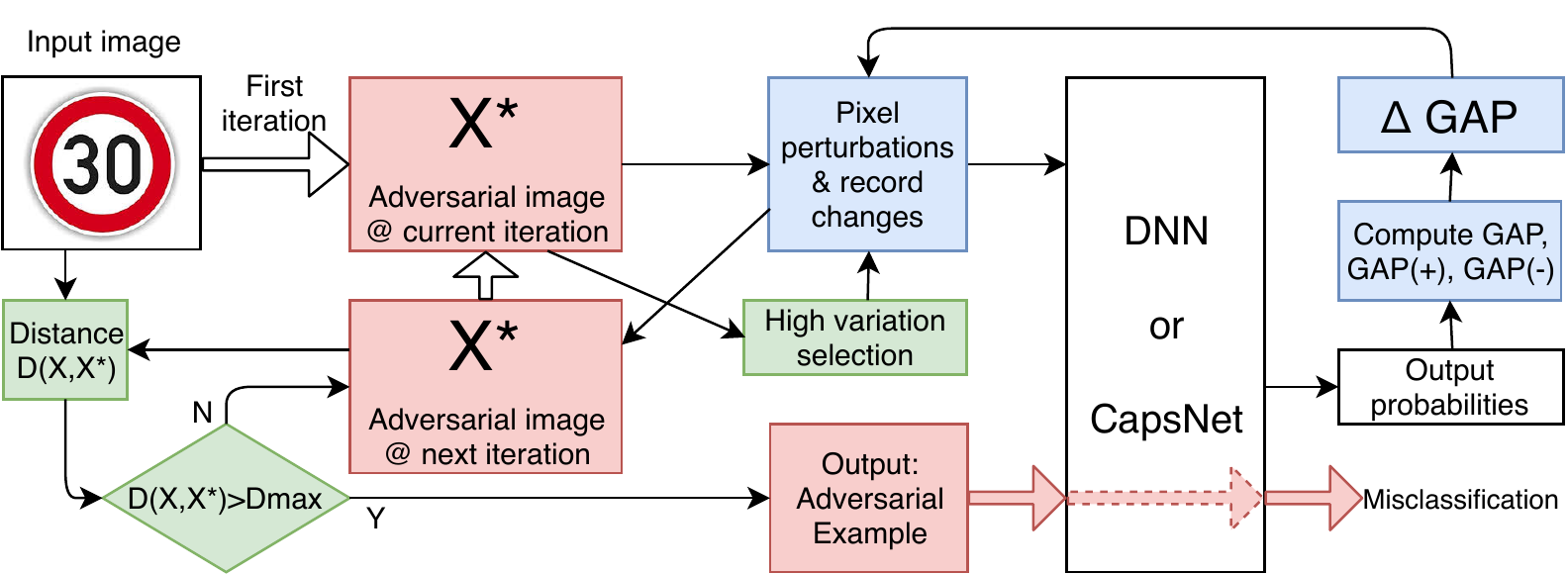}
\vspace*{0mm}
\caption{Our algorithm to generate adversarial examples. The blue-colored boxes are aimed to fool the network, while the green-colored boxes control the imperceptibility of the adversarial attack, whose images at the various stages are represented in the red boxes.}
\label{fig:capsattack}
\end{figure}

\begin{figure}[t]
\begin{algorithm}[H]
\caption{\textbf{: Our Algorithm for Generating Adversarial Attacks}}\label{Greedy algorithm}
\begin{algorithmic}
 \STATE Given: original sample X, maximum human perceptual distance $D_{MAX}$, noise magnitude $\delta$, $M \cdot N$ pixels, target class, P, V
\WHILE{$D(X^*,X)<D_{MAX}$}
\STATE -Compute \textit{Standard Deviation SD} for every pixel
\STATE -Select a subset P of pixels included in the area of $M \cdot N$ pixels with the highest \textit{SD} for every channel
\STATE -Compute $GAP$, $GAP(-)$, $GAP(+)$
\IF{$GAP(-) > GAP(+)$}
\STATE $VariationPriority(x_{i,j})=[GAP(-)-GAP] \cdot SD(x_{i,j})$
\ELSE
\STATE $VariationPriority(x_{i,j})=[GAP(+) - GAP] \cdot SD(x_{i,j})$
\ENDIF
\STATE -Sort in descending order VariationPriority for every channel
\STATE -Select V pixels with highest VariationPriority between the three channels
\IF{$GAP(-) > GAP(+)$}
\STATE Subtract noise with magnitude $\delta$ from the pixel in the respective channel
\ELSE
\STATE Add noise with magnitude $\delta$ to the pixel in the respective channel
\ENDIF
\STATE -Compute $D(X^*,X)$ as the sum of the $D(X^*,X)$ of every channel
\STATE -Update the original example with the adversarial one
\ENDWHILE
\end{algorithmic}
\end{algorithm}
\end{figure}

We propose an iterative algorithm that automatically generates targeted imperceptible and robust adversarial examples in a black-box scenario, i.e., we assume that the attacker has access to the input image and to the output probabilities vector, but not to the network model. Our algorithm is shown in Figure \ref{fig:capsattack} and Algorithm \ref{Greedy algorithm}. The goal of our iterative algorithm is to modify the input image to maximize the gap function (\textit{imperceptibility}) until the distance between the original and the adversarial example is under $D_{MAX}$ (\textit{robustness}). The algorithm takes in account the fact that every pixel is composed of three different values, since the images are based on three channels. Compared to the algorithm proposed by Luo et al. (2018), our attack is applied to a set of pixels with the highest standard deviation at every iteration in order to create imperceptible perturbations. Moreover, our algorithm automatically decides if it is more effective to add or subtract the noise, to maximize the gap function, according to the values of two parameters, $GAP(+)$ and $GAP(-)$. These modifications increase the imperceptibility and the robustness of the attack. For the sake of clarity, we also have expressed the formula used to compute the standard deviation in a more comprehensive form.

Our algorithm is composed of the following steps:
\begin{itemize}[leftmargin=*]
\vspace*{-2mm}
    \item Select a subset $P$ of pixels, included in the area of $M \cdot N$ pixels, with the highest SD for every channel\footnote{The images have 3 channels, since they are in RGB format.}, so that their possible modification is difficult to detect.
    \vspace*{-1mm}
    \item Compute the gap function as the difference between the probability of the target class, chosen as the class with the second highest probability, and the maximum output probability.
    \vspace*{-1mm}
    \item For each pixel of P, compute $GAP(+)$ and $GAP(-)$: these quantities correspond to the values of the gap function, estimated by adding and by subtracting, respectively, a perturbation unit to each pixel. These gaps are useful to decide if it is more effective to add or subtract the noise. For each pixel of P, we consider the greatest value between $GAP(+)$ and $GAP(-)$ to maximize the distance between the two probabilities.
    \vspace*{-1mm}
    \item For each pixel of $P$, calculate the Variation Priority by multiplying the gap difference to the SD of the pixel. This quantity indicates the efficacy of the pixel perturbation.
    \vspace*{-1mm}
    \item For every channel, the $P$ values of Variation Priority are ordered and the highest V values, between the three channels, are perturbed.
    \vspace*{-1mm}
    \item Starting from $3 \cdot P$ values of Variation Priority, only V values are perturbed. According to the highest value of the previous computed $GAP(+)$ and $GAP(-)$, the noise is added or subtracted.
    \vspace*{-1mm}
    \item Once the original input image is replaced by the adversarial one, the next iteration starts. The iterations continue until the distance $D$ overcomes $D_{MAX}$.
    \vspace*{-1mm}
\end{itemize}

\section{Evaluating our Attack algorithm}
\subsection{Experimental Setup}

We apply our algorithm, showed in Section 3.2, to the previously described CapsuleNet, LeNet and VGGNet. To verify how our algorithm works, we test it on two different examples. We consider M=N=32, because the GTRSB dataset is composed of 32$\cdot$32 images, P=100 and V=20. The value of $\delta$ is equal to the 10\% of the maximum value between all the pixels. The parameter $D_{MAX}$ depends on the SD of the pixels of the input image: its value changes according to the different examples because $D(X^*,X)$ does not increase in the same way for each example.

\subsection{Our algorithm applied to the CapsuleNet}
\label{subsec:attack_caps}

\begin{figure}[t]
\centering
\begin{minipage}[t]{.13\linewidth}
\subfloat[]{
\includegraphics[width=1.3\linewidth]{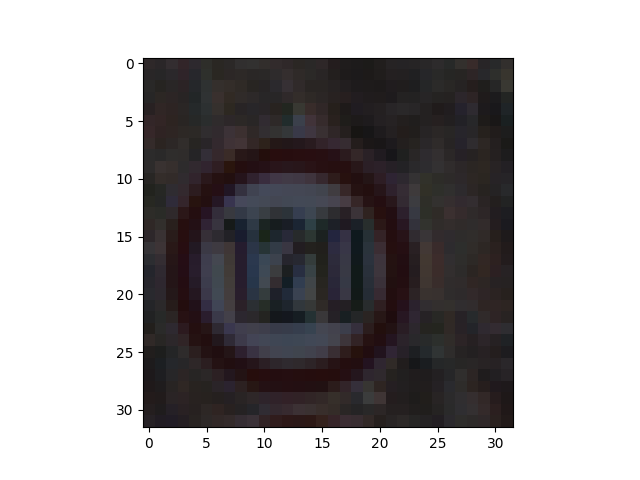}
\label{fig:49_original}}
\end{minipage}
\hfill
\begin{minipage}[t]{.13\linewidth}
\subfloat[]{
\includegraphics[width=1.3\linewidth]{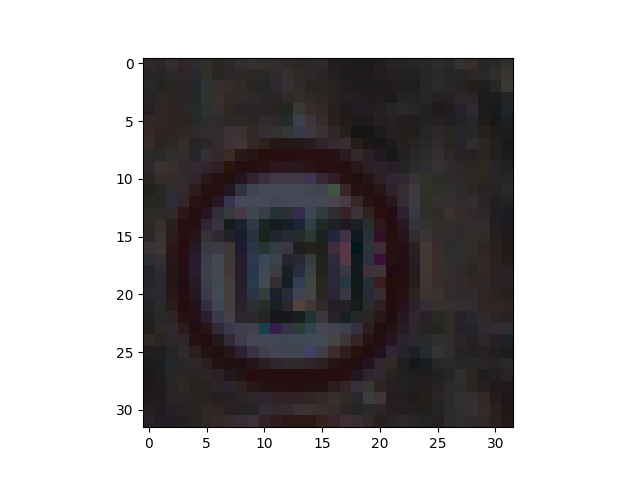}
\label{fig:49_corrtc_it15}}
\end{minipage}
\hfill
\begin{minipage}[t]{.13\linewidth}
\subfloat[]{
\includegraphics[width=1.3\linewidth]{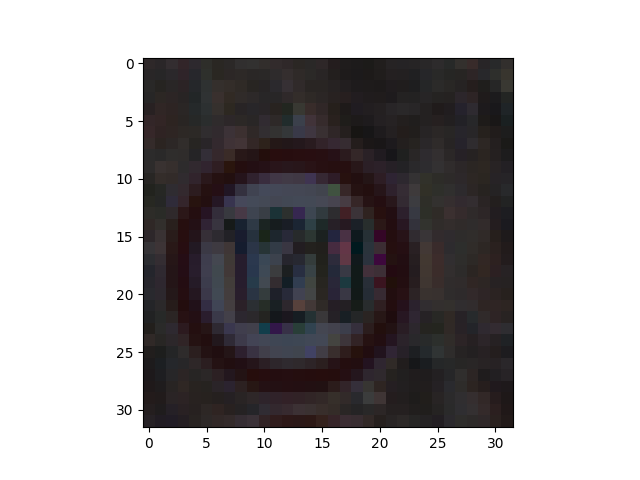}
\label{fig:49_corrtc_it20}}
\end{minipage}
\hfill
\begin{minipage}[t]{.13\linewidth}
\subfloat[]{
\includegraphics[width=1.3\linewidth]{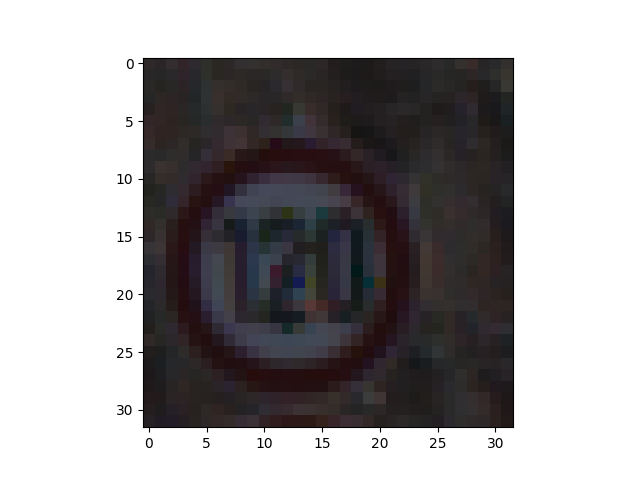}
\label{fig:49_tc30_it25}}
\end{minipage}
\hfill
\begin{minipage}[t]{.13\linewidth}
\subfloat[]{
\includegraphics[width=1.3\linewidth]{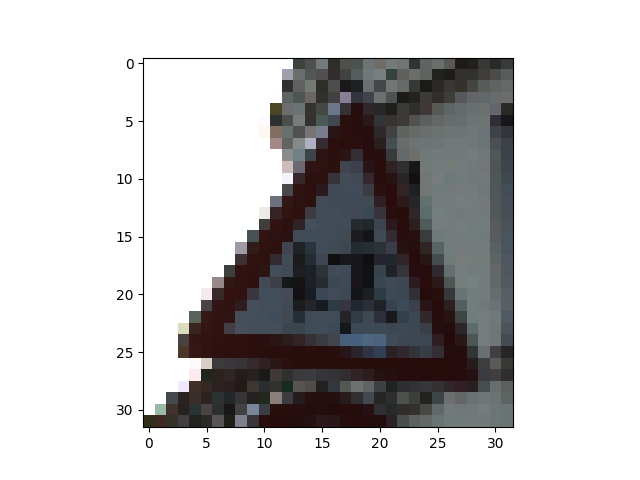}
\label{fig:3421_original}}
\end{minipage}
\hfill
\begin{minipage}[t]{.13\linewidth}
\subfloat[]{
\includegraphics[width=1.3\linewidth]{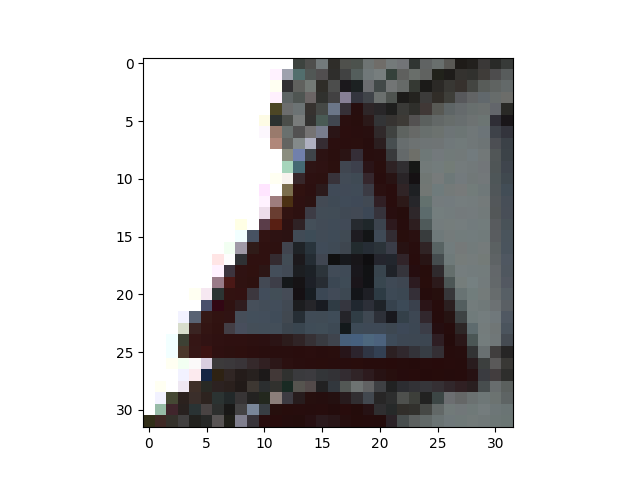}
\label{fig:3421_it16}}
\end{minipage}
\hfill
\begin{minipage}[t]{.13\linewidth}
\subfloat[]{
\includegraphics[width=1.3\linewidth]{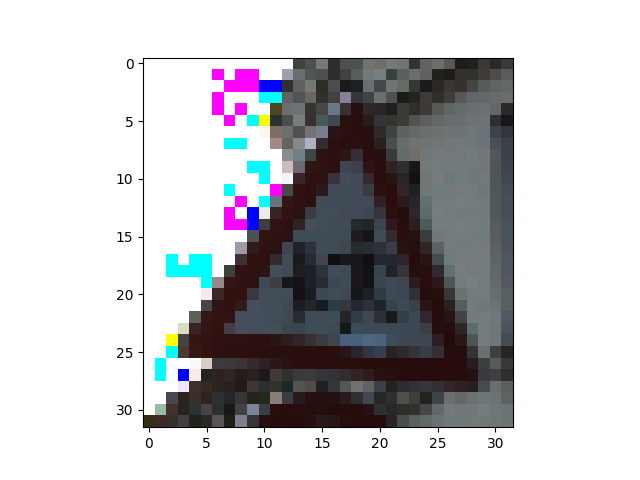}
\label{fig:3421_it33}}
\end{minipage}
\caption{Images for the attack applied to the CapsuleNet: (a) Original input image of Example 1. (b) Image misclassified by the CapsuleNet at the iteration 13 for Case I. (c) Image misclassified by the CapsuleNet at the iteration 16 for Case I. (d) Image at the iteration 12 for Case II. (e) Original input image for Example 2.\\(f) Image at the iteration 5, applied to the CapsuleNet. (g) Image misclassified by the CapsuleNet at iteration 21.}
\label{fig:49caps}
\end{figure}

We test the CapsuleNet on two different examples, shown in Figures \ref{fig:49_original} (Example 1) and \ref{fig:3421_original} (Example 2). For the first one, we distinguish two cases, in order to test whether our algorithm works independently from the target class, and if the final results are different: 
\begin{itemize}[leftmargin=*]
\vspace*{-2mm}
    \item \textbf{Case I:} the target class is the class relative to the second highest probability between all the initial output probabilities.
    \vspace*{-1mm}
    \item \textbf{Case II:} the target class is the class relative to the fifth highest probability between all the initial output probabilities.
    \vspace*{-2mm}
\end{itemize}

\begin{figure}[t]
\centering
\begin{minipage}[t]{.32\linewidth}
\subfloat[]{
\includegraphics[width=\linewidth]{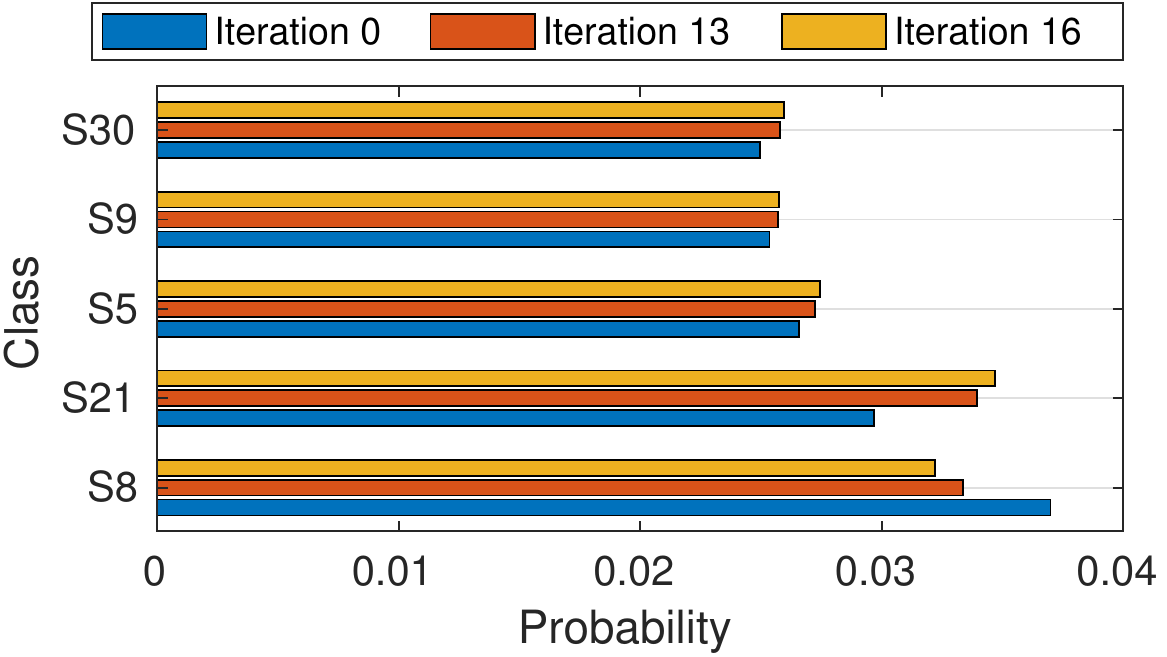}
\label{fig:histo_im49_caps_tc_correct}}
\end{minipage}
\hfill
\begin{minipage}[t]{.32\linewidth}
\subfloat[]{
\includegraphics[width=\linewidth]{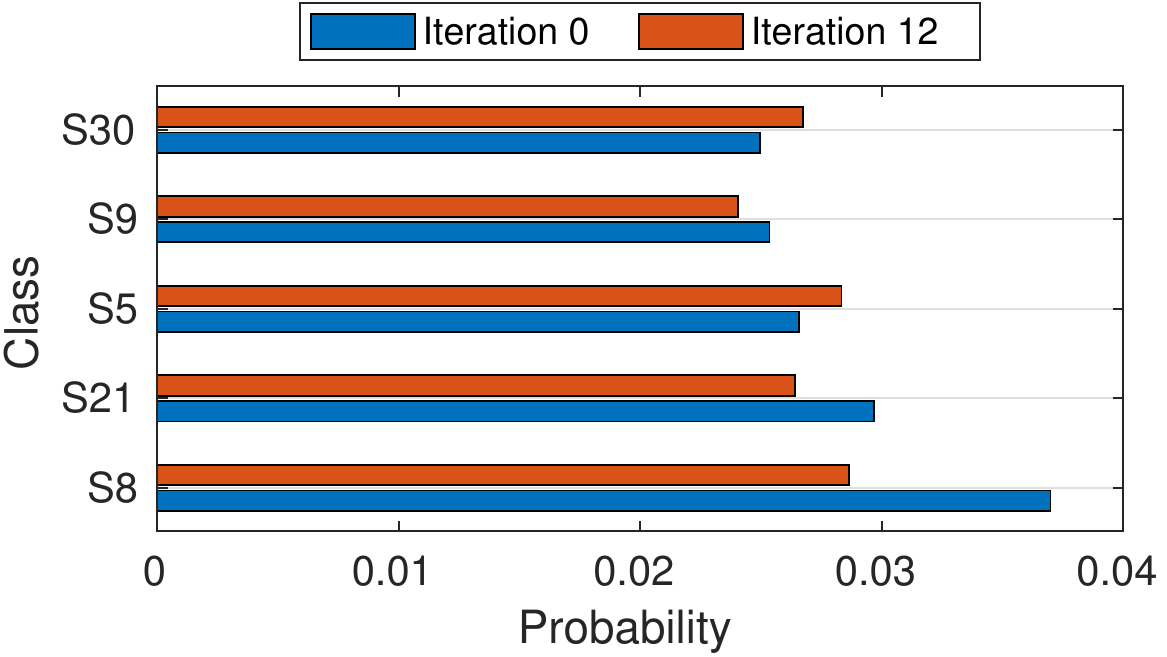}
\label{fig:histo_im49_caps_tc_30}}
\end{minipage}
\hfill
\begin{minipage}[t]{.32\linewidth}
\subfloat[]{
\includegraphics[width=\linewidth]{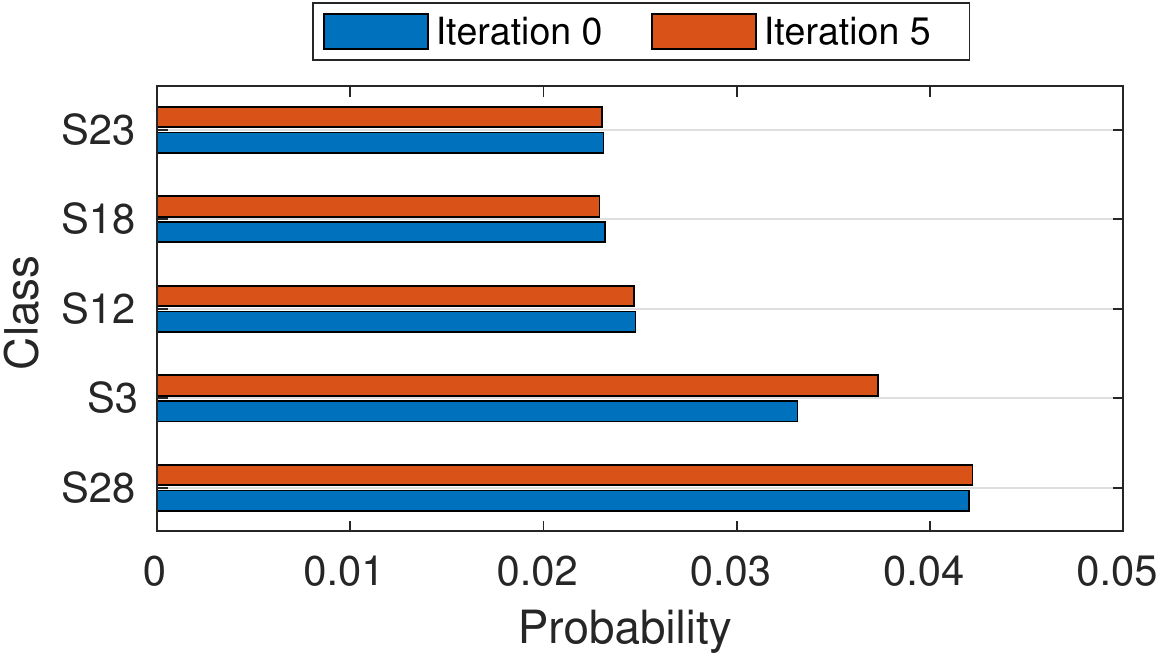}
\label{fig:histo_im3421_correcttc}}
\end{minipage}
\hspace{.03\linewidth}
\caption{CapsuleNet results: (a) Output probabilities of the Example 1 - Case I: blue bars represent the starting probabilities, orange bars the probabilities at the point of misclassification and yellow bars at the $D_{MAX}$.\\(b) Output probabilities of the Example 1 - Case II: blue bars represent the starting probabilities and orange bars the probabilities at the $D_{MAX}$. (c) Output probabilities of the Example 2: blue bars represent the starting probabilities, and orange bars the probabilities at the $D_{MAX}$.}
\label{fig:histo_caps}
\end{figure}

By analyzing the examples in Case I and Case II, we can observe that:

\begin{enumerate}[leftmargin=*]
\vspace*{-2mm}
       \item The CapsuleNet classifies the input image shown in Figure \ref{fig:49_original} as ``120 km/h speed limit'' (S8) with a probability equal to 0.0370.
       
       For the Case I, the target class is ``Double curve'' (S21) with a probability equal to 0.0297. After 13 iterations of our algorithm, the image (in Figure \ref{fig:49_corrtc_it15}) is classified as ``Double curve'' with a probability equal to 0.0339. Hence, \textit{the probability of the target class has overcome the initial one}, as shown in Figure \ref{fig:histo_im49_caps_tc_correct}. At this step, the distance $D(X^*,X)$ is equal to 434.20. Increasing the number of iterations, the robustness of the attack increases as well, because the gap between the two probabilities increases, but also the perceptibility of the noise becomes more evident. After the iteration 16, the distance grows above $D_{MAX}=520$: the sample is represented in Figure \ref{fig:49_corrtc_it20}.
    
    For the Case II, the probability relative to the target class ``Beware of ice/snow'' (S30) is equal to 0.0249, as shown in Figure \ref{fig:histo_im49_caps_tc_30}. The gap between the maximum probability and the probability of the target class is higher than the gap in Case I. After 12 iterations, the network has not misclassified the image yet (see Figure \ref{fig:49_tc30_it25}). In Figure \ref{fig:histo_im49_caps_tc_30} we can observe that the gap between the two classes has decreased, but not enough for a misclassification. However, at this iteration, the value of the distance overcomes $D_{MAX}=520$. In this case, we show that our algorithm would need more iterations to misclassify, at the cost of more perceivable perturbations.
    \vspace*{-1mm}
      \item The CapsuleNet classifies the input image shown in Figure \ref{fig:3421_original} as ``Children crossing'' (S28) with a probability equal to 0.042. The target class is ``60 km/h speed limit'' (S3) with a probability equal to 0.0331. After 5 iterations, the distance overcomes $D_{MAX}=250$, while the network has not misclassified the image yet (see Figure \ref{fig:3421_it16}), because the probability of the target class does not overcome the initial maximum probability, as shown in Figure \ref{fig:histo_im3421_correcttc}. The misclassification appears at the iteration 21, when the sample (see Figure \ref{fig:3421_it33}). However, the perturbation is very percievable.  
    \vspace*{-2mm}
\end{enumerate}

\subsection{Our algorithm applied to a 9-layer VGGNet and a 5-layer LeNet}
\label{subsec:vgg}

\begin{figure}[t]
\centering
\begin{minipage}[t]{.13\linewidth}
\subfloat[]{
\includegraphics[width=1.3\linewidth]{49_original.png}
\label{fig:49_original_vgg}}
\end{minipage}
\hfill
\begin{minipage}[t]{.13\linewidth}
\subfloat[]{
\includegraphics[width=1.3\linewidth]{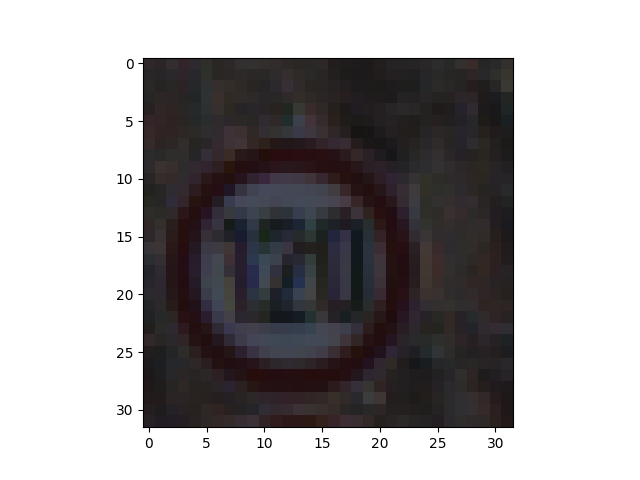}
\label{fig:49vgg_it4}}
\end{minipage}
\hfill
\begin{minipage}[t]{.13\linewidth}
\subfloat[]{
\centering
\includegraphics[width=1.3\linewidth]{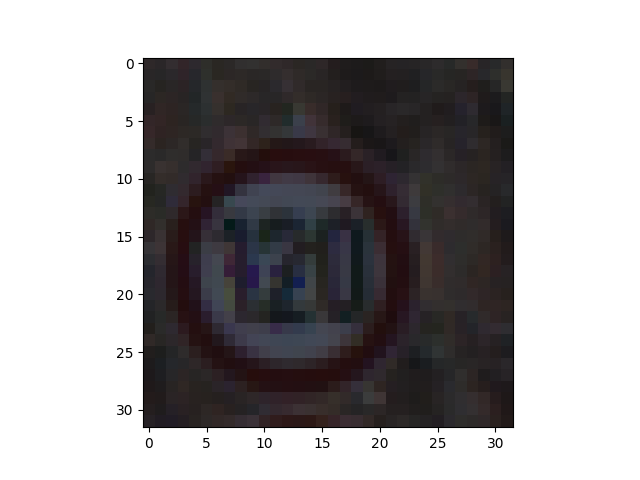}
\label{fig:49_vgg_it15}}
\end{minipage}
\hfill
\begin{minipage}[t]{.13\linewidth}
\subfloat[]{
\includegraphics[width=1.3\linewidth]{3421_original.png}
\label{fig:3421_original_vgg}}
\end{minipage}
\hfill
\begin{minipage}[t]{.13\linewidth}
\subfloat[]{
\includegraphics[width=1.3\linewidth]{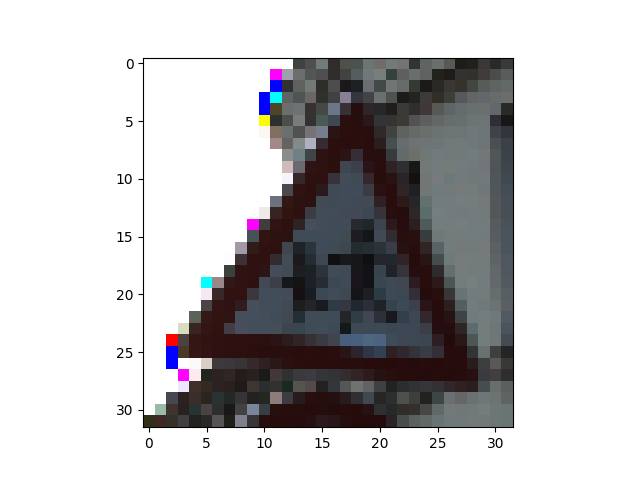}
\label{fig:3421_it9_vgg}}
\end{minipage}
\hfill
\begin{minipage}[t]{.13\linewidth}
\subfloat[]{
\includegraphics[width=1.3\linewidth]{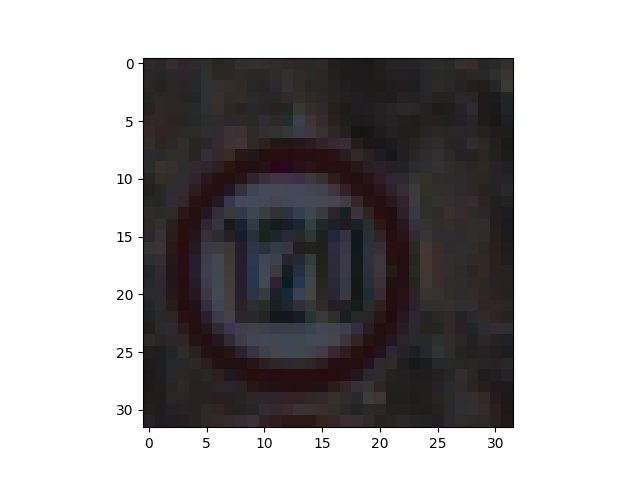}
\label{fig:lenet0}}
\end{minipage}
\hfill
\begin{minipage}[t]{.13\linewidth}
\subfloat[]{
\includegraphics[width=1.3\linewidth]{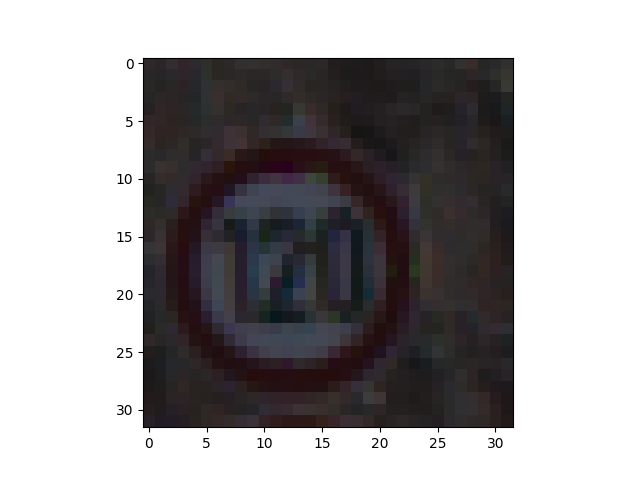}
\label{fig:lenet11}}
\end{minipage}
\hspace{.03\linewidth}
\caption{Images for the attack applied to the CNNs: (a) Original input image for Example 1 (b) Image at the iteration 3, applied to the VGGNet. (c) Image at the iteration 9, misclassified by the VGGNet. (d) Original input image for Example 2. (e) Image at the iteration 2, applied to the VGGNet. (a) Image at the iteration 6, misclassified by the LeNet.. (b) Image at the iteration 13, misclassified by the LeNet.}
\label{fig:49vgg}
\end{figure}

\begin{figure}[t]
\centering
\begin{minipage}[t]{.32\linewidth}
\subfloat[]{
\includegraphics[width=\linewidth]{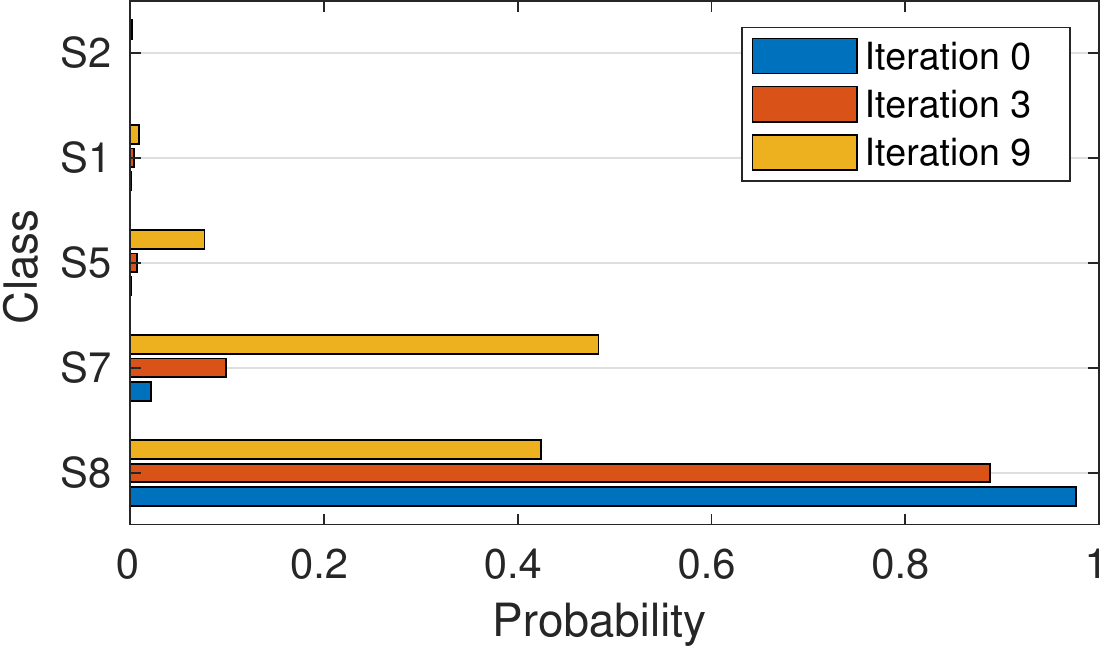}
\label{fig:histo_im49_vgg_tc_correct}}
\end{minipage}
\hfill
\begin{minipage}[t]{.32\linewidth}
\subfloat[]{
\includegraphics[width=\linewidth]{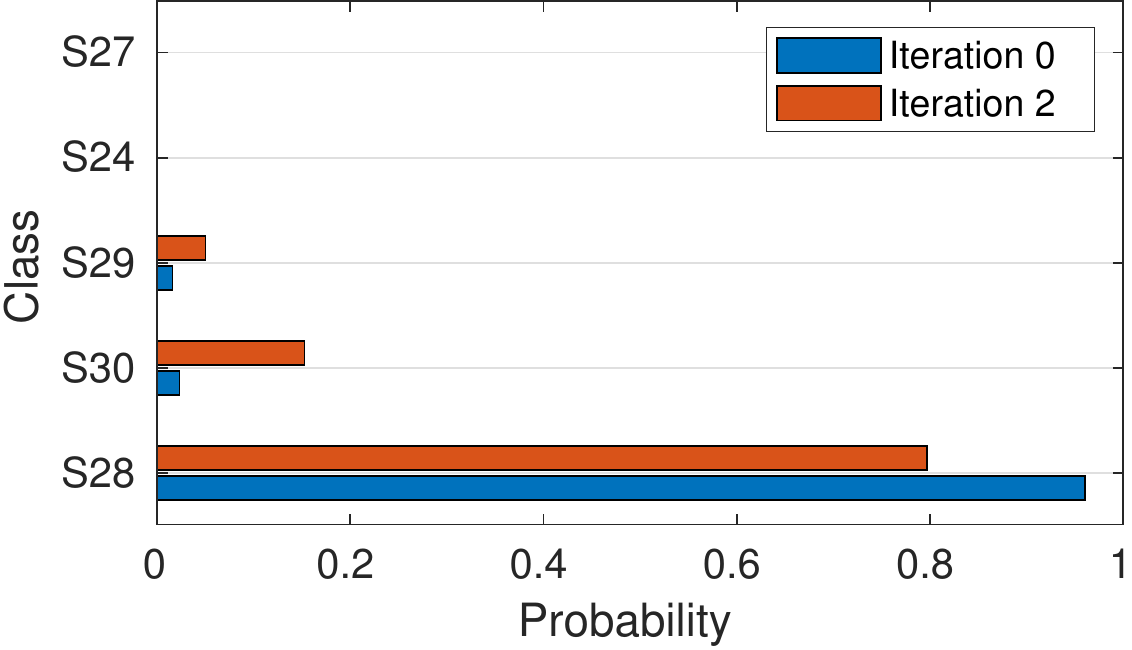}
\label{fig:histo_im3421_vgg}}
\end{minipage}
\hfill
\begin{minipage}[t]{.32\linewidth}
\subfloat[]{
\includegraphics[width=\linewidth]{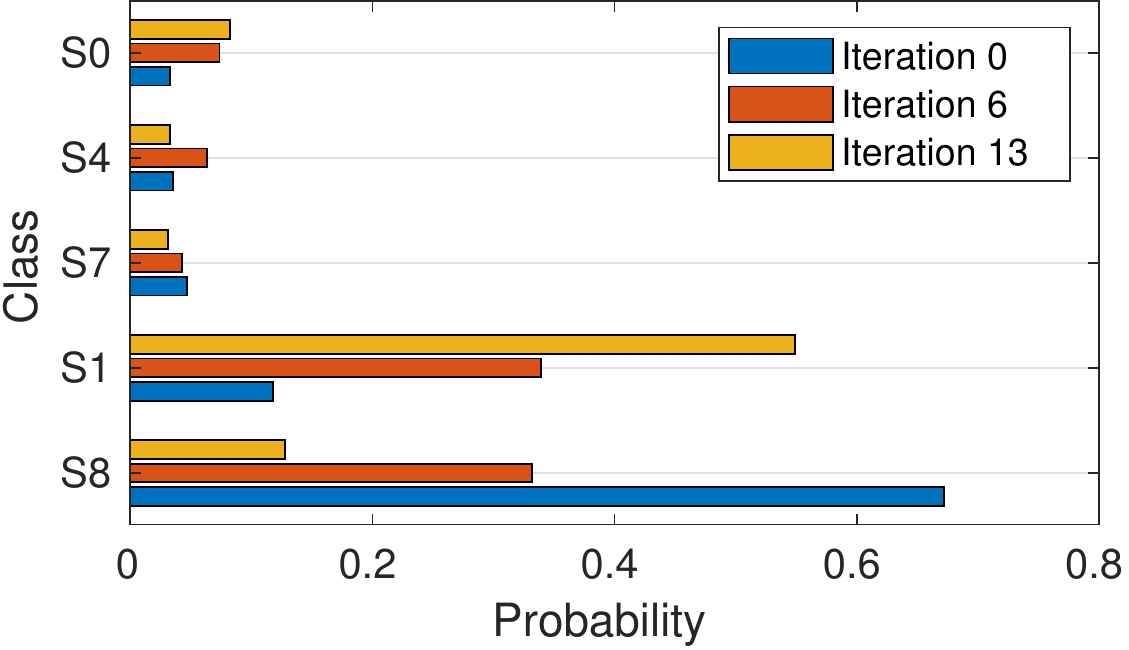}
\label{fig:histo_im49_lenet}}
\end{minipage}
\vspace*{0mm}
\caption{CNNs results. (a) Output probabilities of the Example 1 on the VGGNet: blue bars represent the starting probabilities, orange bars the probabilities at the point of misclassification and yellow bars at the $D_{MAX}$. (b) Output probabilities of the Example 2 on the VGGNet: blue bars represent the starting probabilities and orange bars the probabilities at the $D_{MAX}$. (c) Output probabilities of the Example 1 on the LeNet: blue bars represent the starting probabilities, orange bars the probabilities at the point of misclassification and yellow bars at the $D_{MAX}$.}
\label{fig:vgg_histo}
\vspace*{0mm}
\end{figure}

To compare the robustness of the CapsuleNet and the 9-layer VGGNet, we choose to evaluate the previous two examples, which have been applied to the CapsuleNet. For the Example 1, we consider only the Case I as benchmark. The VGGNet classifies the input images with different output probabilities, compared to the ones obtained by the CapsuleNet. Therefore, our metric to evaluate how much the VGGNet is resistant to our attack is based on the value of the gap at the same distance.

To compare the robustness of the CapsuleNet and the 5-layer LeNet, we only consider the Example 1, because the Example 2 is already classified incorrectly by the LeNet\footnote{Note: the image of Example 2 belongs to the subset of images that are correctly classified by the CapsuleNet and the VGGNet, but incorrectly by the LeNet}. Applying our algorithm to the LeNet, we observe that, as expected, it is more vulnerable than the CapsuleNet and the VGGNet.

We can make the following considerations on our examples:
\begin{enumerate}[leftmargin=*]
\vspace*{-2mm}
    \item The VGGNet classifies the input image (in Figure \ref{fig:49_original_vgg}) as ``120 km/h speed limit'' (S8) with a probability equal to 0.976. The target class is ``100 km/h speed limit'' (S7) with a probability equal to 0.021. After 3 iterations, the distance overcomes $D_{MAX}=520$, while the VGGNet has not misclassified the image yet (see Figure \ref{fig:49vgg_it4}) yet: our algorithm would need to perform more iterations before fooling the VGGNet, since the two initial probabilities were very distant, as shown in Figure \ref{fig:histo_im49_vgg_tc_correct}. Such scenario appears after 9 iterations (in Figure \ref{fig:49_vgg_it15}), where the probability of the target class is 0.483.
    \vspace*{-1mm}
    \item The VGGNet classifies the input image (in Figure \ref{fig:3421_original_vgg}) as ``Children crossing'' (S28) with a probability equal to 0.96. The target class is ``Beware of ice/snow'' (S30) with a probability equal to 0.023. After 2 iterations, the distance overcomes $D_{MAX}=250$, while the VGGNet has not misclassified the image yet (see Figure \ref{fig:3421_it9_vgg}). As in the previous case, this scenario is due to the high distance between the initial probabilities, as shown in Figure \ref{fig:histo_im3421_vgg}. We can also notice that the VGGNet reaches $D_{MAX}$ in a lower number of iterations, as compared to the CapsuleNet.
    \vspace*{-1mm}
    \item The LeNet classifies the input image (in Figure \ref{fig:49_original_vgg}) as ``120 km/h speed limit'' (S8) with a probability equal to 0.672. The target class is ``30 km/h speed limit'' (S1) with a probability equal to 0.178. After 6 iterations, the perturbations fool the LeNet, because the image (In Figure \ref{fig:lenet0}), is classified as the target class with a probability equal to 0.339. The perturbations become perceptible after 13 iterations (Figure \ref{fig:lenet11}), where the distance overcomes $D_{MAX}=520$.
    \vspace*{-2mm}
\end{enumerate}

\begin{figure}[t]
\centering
\begin{minipage}[t]{.47\linewidth}
\subfloat[]{
\includegraphics[width=\linewidth]{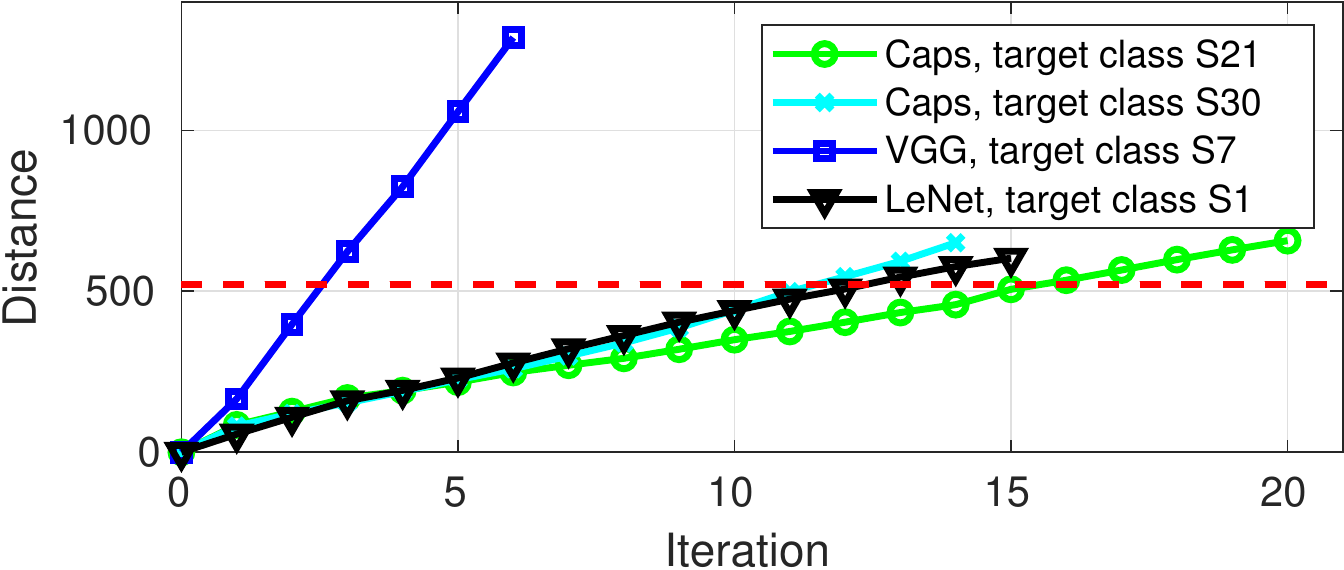}
\label{fig:distance49}}
\end{minipage}
\hfill\begin{minipage}[t]{.47\linewidth}
\subfloat[]{
\includegraphics[width=\linewidth]{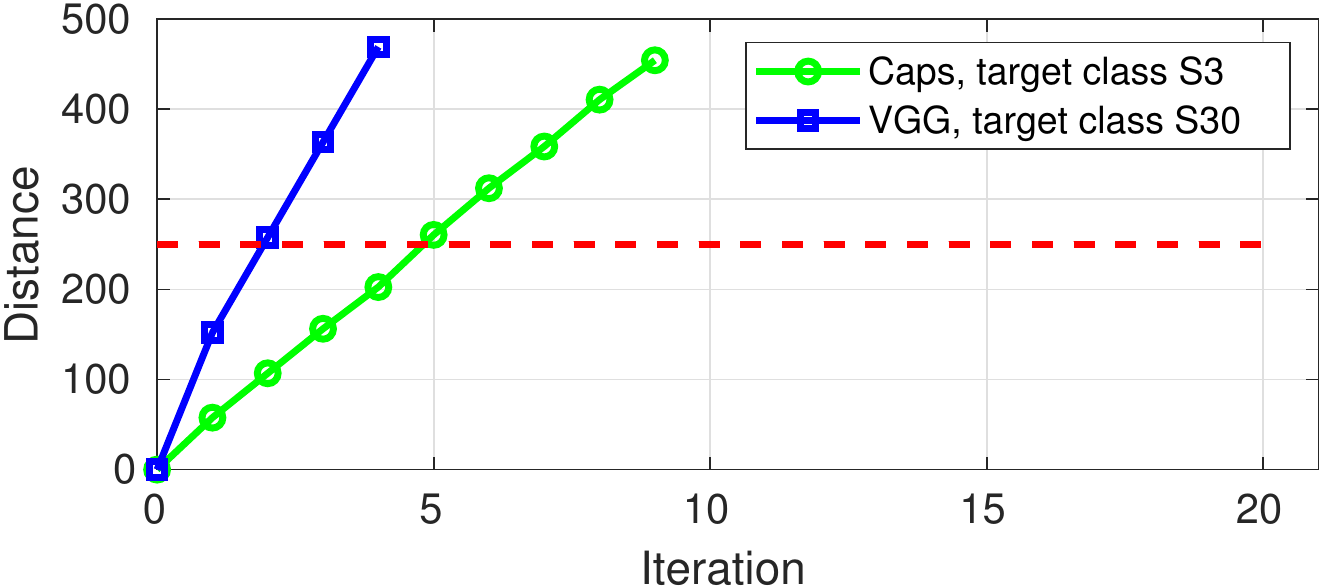}
\label{fig:distance3421}}
\end{minipage}
\caption{ (a) $D(X^*,X)$ behavior for the Example 1. (b) $D(X^*,X)$ behavior for the Example 2.}
\label{fig:distance}
\vspace*{0mm}
\end{figure}

\subsection{Comparison and results}

From our analyses, we can observe that the vulnerability of the 9-layer VGGNet to our adversarial attack is slightly lower than the vulnerability of the CapsuleNet, since the former one requires more perceivable perturbations to be fooled. Our observation is remarked by the graphs in Figure \ref{fig:distance}: the value of $D(X^*,X)$ increases more sharply for the VGGNet than for the CapsuleNet. Hence, the perceivability of the noise in the image can be measured as the value of $D(X^*,X)$ divided by the number of iterations: the noise in the VGGNet becomes perceivable after few iterations. Moreover, we can observe that the choice of the target class plays a key role for the success of the attack.

We notice that other features that evidence the differences between the VGGNet and the CapsuleNet are crucial. The VGGNet is deeper and contains a larger number of weights, while the CapsuleNet can achieve a similar accuracy with a smaller footprint. This effect causes a disparity in the prediction confidence between the two networks. It is clear that the CapsuleNet has a much higher learning capability, compared to the VGGNet, but this phenomena does not reflect to the prediction confidence. Indeed, comparing Figures \ref{fig:histo_caps} and \ref{fig:vgg_histo}, we can notice that the output probabilities predicted by the CapsuleNet are close to each other, even more than the LeNet. However, the perturbations does not affect the output probabilities of the CapsuleNet as much as for the cases of the CNNs. The LeNet, as expected, even though it has a similar depth and similar number of parameters, is more vulnerable than the CapsuleNet.

Moreover, recalling from Section \ref{sec:affine_transformation}, we observed that the VGGNet and the CapsuleNet are more resistant than the LeNet to the affine transformations: this behavior is consistent with the results obtained applying our adversarial attack algorithm.

\section{Conclusions}

In this paper, we proposed an algorithm to generate targeted adversarial attacks in a black box scenario. We applied our attack to the GTSRB dataset and we verified its impact on a CapsuleNet, a 5-layer LeNet and a 9-layer VGGNet. Our experiments show that the CapsuleNet appears more robust than the LeNet to the attack, but slightly less than the VGGNet, because the modifications of the pixels in the traffic signs are less perceivable when our algorithm is applied to the CapsuleNet, rather than to the VGGNet. A serious issue for the CapsuleNet is that the gap between the output probabilities is lower than the one computed on the VGGNet predictions. However, the change in absolute values of the probabilities at the output of the CapsuleNet is lower than their changes in the VGGNet. Hence, further modifications of the CapsuleNet algorithm, aiming to increase the prediction confidence, would be beneficial to improve its robustness.

\end{document}